\pdfoutput=1
\documentclass[11pt]{article}
\usepackage{acl}

\usepackage{times}
\usepackage{latexsym}

\usepackage[T1]{fontenc}
\usepackage[utf8]{inputenc}

\usepackage{microtype}

\usepackage{graphicx}

\newif\ifwithappendix
\withappendixfalse

\newif\ifappendixshown
\appendixshowntrue

\usepackage[T1]{fontenc}
\usepackage[utf8]{inputenc}
\usepackage[textsize=tiny]{todonotes}
\usepackage{graphicx}
\usepackage{booktabs}
\usepackage{latexsym}
\usepackage{microtype}

\usepackage{footnote}
\makesavenoteenv{tabular}
\makesavenoteenv{table}
\makesavenoteenv{table*}

\usepackage{xurl}

\newcommand\minput[1]{%
  \input{#1}%
  \ifhmode\ifnum\lastnodetype=11 \unskip\fi\fi}

\usepackage{hyperref}
\usepackage{xstring}

\newcommand{\noqa}[1]{}
\newcommand{\noqall}[1]{}

\usepackage{amsmath}

\title{
Seamlessly Integrating Tree-Based Positional Embeddings into Transformer Models for Source Code Representation
}

\author{
    \textbf{Patryk Bartkowiak, Filip Graliński} \\
    Adam Mickiewicz University
}

\usepackage{times}
\usepackage{inconsolata}

\begin{document}
\maketitle
\begin{abstract}
Transformer-based models have demonstrated significant success in various source code representation tasks. Nonetheless, traditional positional embeddings employed by these models inadequately capture the hierarchical structure intrinsic to source code, typically represented as Abstract Syntax Trees (ASTs). To address this, we propose a novel tree-based positional embedding approach that explicitly encodes hierarchical relationships derived from ASTs, including node depth and sibling indices. These hierarchical embeddings are integrated into the transformer architecture, specifically enhancing the CodeBERTa model. We thoroughly evaluate our proposed model through masked language modeling (MLM) pretraining and clone detection fine-tuning tasks. Experimental results indicate that our Tree-Enhanced CodeBERTa consistently surpasses the baseline model in terms of loss, accuracy, F1 score, precision, and recall, emphasizing the importance of incorporating explicit structural information into transformer-based representations of source code.
\end{abstract}

\section{Introduction}

Transformer-based models have demonstrated significant advances across numerous source code representation tasks, such as code summarization, clone detection, defect prediction, and semantic search. These models effectively leverage self-attention mechanisms, supplemented by positional embeddings, to encode the sequential order of tokens, thus achieving robust semantic understanding. However, a notable limitation arises from the fundamentally linear positional encodings typically employed by these models, which do not adequately capture the inherently hierarchical nature of programming languages \cite{programs_with_graphs, code_modeling_with_graphs, models_source_code}.

Source code differs markedly from natural language in its explicit, structured hierarchy, most commonly represented as Abstract Syntax Trees (ASTs) \cite{novel_positional_encodings}. An AST encodes critical syntactic and semantic relationships, including nested scopes, parent-child dependencies, and sibling ordering among code constructs. Although sequential positional embeddings, such as sinusoidal or learned encodings \cite{Attention, BERT}, effectively capture linear token sequences, they disregard hierarchical structure entirely. Consequently, current Transformer architectures may overlook important syntactic relationships, potentially limiting performance and generalization capabilities in source code understanding tasks.

Addressing this gap, we propose integrating tree-based positional embeddings into Transformer models, explicitly encoding the hierarchical structure of source code. Our approach introduces embeddings based on token depth within the AST hierarchy and their sibling positions, effectively guiding the self-attention mechanism to recognize and utilize structural context alongside semantic content.

In this paper, we make the following contributions.
\begin{itemize}
\item We propose a novel hierarchical embedding strategy to explicitly encode structural information from ASTs into Transformer-based models.
\item We demonstrate how tree-based positional embeddings can be seamlessly integrated into existing Transformer architectures, specifically CodeBERTa, without significantly increasing the complexity of the model.
\item Through extensive experiments, including masked language modeling (MLM) and code clone detection tasks, we illustrate that our Tree-Enhanced CodeBERTa outperforms similarly sized models in smaller-scale evaluations, highlighting the benefits of incorporating explicit structural context.
\item We provide qualitative analysis using visualization techniques (e.g.\ t-SNE) to demonstrate how tree-based embeddings result in structurally coherent code representations, further validating our theoretical insights.
\end{itemize}

Our findings underscore the importance of integrating hierarchical structural information into Transformer architectures, not only enhancing source code representation but also potentially improving models for broader structured data.

\section{Related Work}

Transformer-based models, including GPT \cite{GPT}, BERT \cite{BERT}, and RoBERTa \cite{RoBERTa}, have significantly advanced natural language processing (NLP) and have subsequently been adapted for source code understanding tasks. This section reviews the relevant literature, classified into transformer models for source code, positional embedding methods, and tree-based code representations.

\paragraph{Transformers for Source Code.}

Pre-trained Transformer models such as CodeBERT and CodeBERTa utilize broad datasets such as CodeSearchNet to effectively learn token-level semantic representations \cite{code_summarization, feng2020codebert, husain_codesearchnet_2019}. Although these models have achieved strong results, their effectiveness stems primarily from capturing semantic relationships between tokens without explicitly modeling the hierarchical structural relationships encoded in Abstract Syntax Trees (ASTs). Consequently, structural nuances, crucial for tasks that require deeper syntactic and semantic understanding, remain largely unaddressed.

\paragraph{Positional Embeddings in Transformers.}

The original Transformer model \cite{Attention} employs sinusoidal positional embeddings to encode token positions within sequences, establishing order-aware representations. Subsequent developments introduced learned positional embeddings, enabling dynamic adaptation during training \cite{relative_position_representations}. Recently, Rotary Positional Embeddings (RoPE) \cite{RoFormer} have emerged as an effective technique for capturing relative positional information in Transformers, achieving superior generalization in sequence modeling tasks. Despite these advancements, positional embeddings typically remain confined to linear positional encodings, which do not inherently capture hierarchical or structural relationships essential in structured data like source code \cite{relative_position_representations, train_short_test_long}.

\paragraph{Tree-Based Representations in Code Analysis.}

Various architectures have leveraged explicit tree-based structures for representing code, particularly AST-based models such as Tree-LSTMs \cite{TreeLSTM}, graph neural networks (GNNs) \cite{tree_structured_LSTM, amr_parsing}, and specialized code-generation models like code2seq \cite{code2seq} and TreeGen \cite{TreeGen}. These models utilize tree structures to enhance code comprehension, program synthesis, and code summarization by explicitly encoding structural information. However, these methods generally rely on specialized architectures, often incompatible or challenging to integrate directly with the standard Transformer architecture. Consequently, practical adaptation in Transformer-based code models has been limited.

Recent approaches have explicitly introduced AST information into positional embeddings within Transformer architectures. Peng et al. (2022) propose a Tree-Transformer that encodes each AST node's position using a two-dimensional coordinate scheme (sibling index and parent's child count), injecting both local and global structural biases \cite{rethinking}. Similarly, Oh and Yoo (2024) introduce CSA-Trans, which utilizes a dedicated \emph{Code Structure Embedder} to learn structure-aware positional embeddings through a disentangled attention mechanism \cite{csa_trans}. In contrast, our Tree-Enhanced CodeBERTa integrates hierarchical positional embeddings directly into an existing pre-trained Transformer (CodeBERTa), explicitly encoding depth and sibling indices. This maintains simplicity and scalability, avoiding the complexity of specialized attention modules or significant architectural alterations.

\paragraph{Our Contribution.}

Unlike prior works that incorporate hierarchical structures into Transformer models through specialized architectures, our approach directly integrates tree-based positional embeddings—encoding depth and sibling indices—into an existing Transformer framework. This allows for richer hierarchical representations without altering the model's core design. We empirically validate its effectiveness in masked language modeling and clone detection, demonstrating measurable improvements in source code representation.

\section{Theoretical Foundations}

\begin{figure}[t]
    \centering
    \includegraphics[width=0.8\columnwidth]{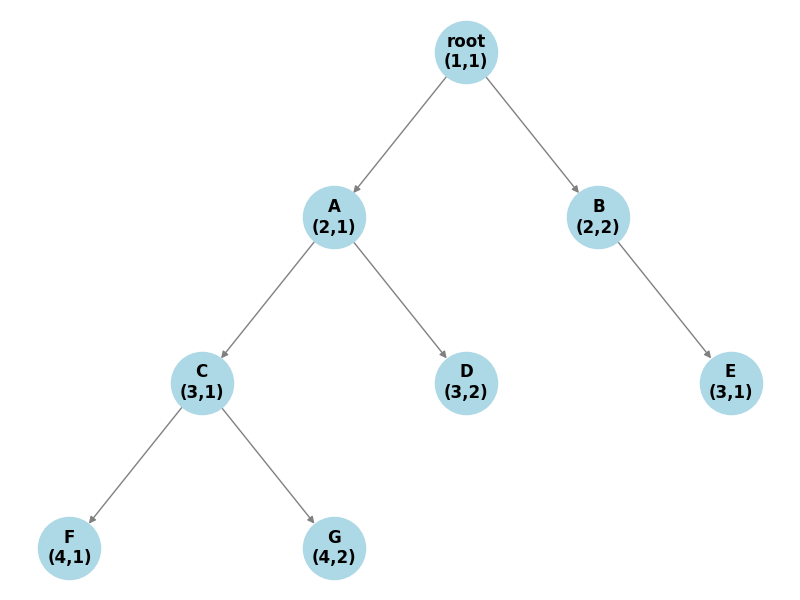}
    \caption{Visualization of hierarchical positional encoding in an AST. Each node is labeled with its name and corresponding hierarchical position \((\text{depth}, \text{sibling index})\), illustrating how depth and sibling relationships are assigned in tree-based positional embeddings.}
    \label{fig:hierarchical-positions}
\end{figure}

Traditional positional embeddings employed in Transformer models typically assume a linear sequence of tokens, effectively capturing the sequential order but failing to represent the complex hierarchical structures inherent in many forms of structured data, notably source code. Source code is commonly represented by Abstract Syntax Trees (ASTs), which explicitly encode syntactic and semantic relationships such as nesting, parent-child dependencies, and sibling ordering. A linear positional encoding is inadequate for capturing such hierarchical nuances, motivating the need for hierarchical positional encodings.

Hierarchical positional embeddings provide a principled approach to representing each node (or token) by encoding its structural position within a tree. Formally, each node's hierarchical position can be represented by a path from the root to that node. Let \( F(x) \) denote the hierarchical position of node \( x \), defined recursively as follows:



\begin{itemize}
    \item \textbf{Base Case}: The root node is assigned a fixed initial position:
    \[
    F(\text{root}) = (1,1).
    \]
    \item \textbf{Recursive Step}: For a non-root node \( x \), its position is determined recursively from its parent \( f(x) \):
    \[
    F(x) = (F(f(x))_1 + 1, i_x),
    \]
    where:
    \begin{itemize}
        \item \( F(f(x))_1 \) refers to the depth component of the parent's position.
        \item \( i_x \) is the index of \( x \) among its siblings.
    \end{itemize}
\end{itemize}

\textbf{Figure~\ref{fig:hierarchical-positions}} provides a visual representation of hierarchical positional encoding, illustrating how depth and sibling indices are assigned to each node in an AST. This structure enables the model to capture both global (depth) and local (sibling order) relationships, which are then transformed into learned embedding vectors.

Each hierarchical position uniquely encodes both local (sibling order) and global (depth in the hierarchy) structural context. These positional values are transformed into learned embedding vectors at each level and then aggregated to form a single positional embedding vector \( P(x) \):

\[
\begin{aligned}
P(x) = \text{Aggregate}\bigl(&h(F(x)_1), h(F(x)_2) \bigr),
\end{aligned}
\]

where \( F(x)_1 \) represents the depth of node \( x \) and \( F(x)_2 \) represents the sibling index \( i_x \), with their corresponding learned embeddings \( h(F(x)_1) \) and \( h(F(x)_2) \).

This formulation allows Transformer models to inherently interpret structural relationships between tokens or nodes. Tokens with similar structural contexts (e.g., siblings or nodes within the same subtree) naturally receive similar positional embeddings, guiding the self-attention mechanism to focus appropriately on structurally relevant elements.

Moreover, hierarchical positional embeddings enhance the model’s capability to capture long-range dependencies inherent in structured data. By explicitly encoding tree positions rather than linear indices, hierarchical positional embeddings facilitate the model's understanding of relationships between tokens that may be distant in a linear sequence yet closely related structurally.

\section{Methodology}

\paragraph{Overview.}

We propose Tree-Enhanced CodeBERTa, an extension of CodeBERTa that integrates tree-based positional embeddings explicitly derived from Abstract Syntax Trees (ASTs). By incorporating \textbf{Depth Embeddings} and \textbf{Sibling Index Embeddings}, our model captures hierarchical relationships inherent in source code, enhancing both syntactic and semantic understanding.

\subsection{Tree-Based Positional Embeddings}

To extract hierarchical structural information, we generate Abstract Syntax Trees (ASTs) from source code using Tree-Sitter \cite{tree_sitter}, a widely used incremental parser supporting multiple programming languages. Tree-Sitter allows efficient parsing and provides structured representations that align well with tokenized inputs, enabling precise mapping of hierarchical relationships.

We introduce two main categories of tree-based positional embeddings:
\begin{itemize}
\item \textbf{Depth Embeddings:} Each token receives an embedding based on its hierarchical depth, represented as \( F(x)_1 \), where deeper nodes correspond to more nested structures such as loops, conditionals, or function bodies.
\item \textbf{Sibling Index Embeddings:} Tokens are embedded based on their relative positions among sibling nodes within the AST, maintaining local ordering essential to understanding structures like function arguments, statements within blocks, and ordered code constructs.
\end{itemize}

Additionally, we introduce a \textbf{Tree Attention Mask}, designed to focus self-attention on structurally related tokens within the AST, thus reducing noisy attention to padding or structurally irrelevant tokens.

\subsection{Integration into Transformer Architecture}

We explore three strategies for embedding integration into the existing CodeBERTa embedding framework:

\textbf{Sum Embeddings}: Structural embeddings (depth and sibling index embeddings) are summed element-wise with standard token embeddings (word embeddings, positional embeddings, and type embeddings), forming a single unified embedding without explicit distinction between semantic and structural contributions.

\textbf{Weighted Sum Embeddings:} A set of learnable weights dynamically balances the contributions of token, depth, sibling index, and positional embeddings. This allows the model to adaptively emphasize the most relevant structural information during training.

\textbf{Concatenation Embeddings:} Structural embeddings (depth and sibling indices) are concatenated with standard token embeddings, followed by a linear projection layer to reduce dimensionality and control model complexity. This method significantly increases the representational power of embeddings but at the cost of higher parameter count and computational complexity.

\begin{figure}[t]
\includegraphics[width=\columnwidth]{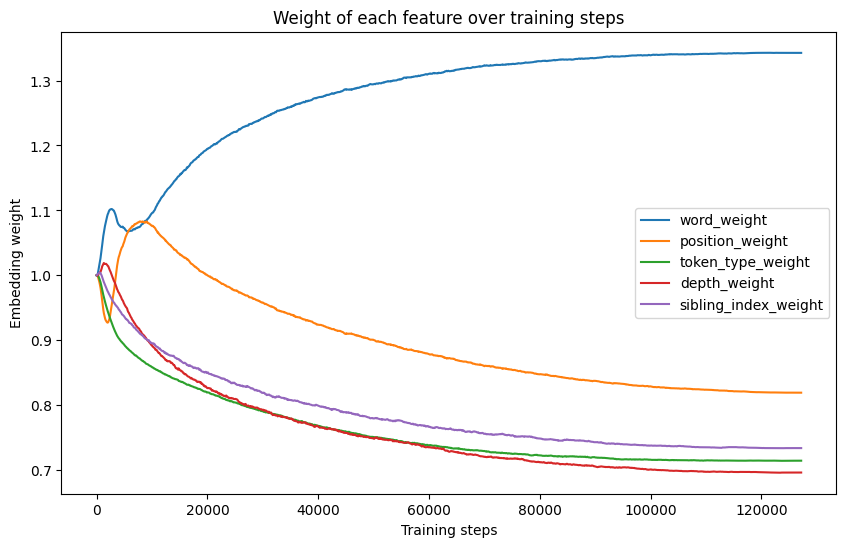}
\caption{Evolution of embedding weights during training for the Weighted Sum configuration. 
The plot shows how different embedding components (word embeddings, positional embeddings, token type embeddings, depth-based embeddings, and sibling index embeddings) are dynamically weighted over training steps. Word embeddings gain prominence, while structural embeddings (depth and sibling index) gradually decrease, indicating their strongest influence early in training.}
\label{fig:embedding-weights}
\end{figure}

The evolution of embedding weights throughout training in the Weighted Sum configuration is illustrated in Figure~\ref{fig:embedding-weights}, demonstrating how the model dynamically adjusts emphasis on structural information.

\subsection{Pretraining and Fine-Tuning Tasks}

To comprehensively evaluate the effectiveness of Tree-Enhanced CodeBERTa, we perform experiments across two core tasks: masked language modeling (MLM) pretraining and clone detection fine-tuning. These tasks are selected to assess the model's ability to leverage hierarchical structural information during both initial representation learning and downstream task adaptation.

\paragraph{Masked Language Modeling (MLM)}

We pretrain the model using the Masked Language Modeling objective on the CodeSearchNet dataset. During pretraining, randomly masked tokens are predicted based on their surrounding context, enriched with hierarchical positional embeddings. This setup assesses the capability of our proposed embeddings to capture both local and global structural relationships inherent in source code.

\paragraph{Clone Detection}

We fine-tune the pretrained model on a clone detection dataset \cite{PoolC} comprising approximately 600,000 code snippet pairs, classifying pairs as functionally equivalent (clones) or distinct. The objective of this task is to measure the practical advantages of hierarchical embeddings in discriminating structurally similar but semantically distinct code snippets, reflecting real-world benefits for source code understanding tasks.

\paragraph{Experimental Setup}

Our Tree-Enhanced CodeBERTa model is based on the CodeBERTa-small architecture, a 6-layer Transformer with 83.5 million parameters. It follows a RoBERTa-like architecture with the same number of layers and attention heads as DistilBERT. While the backbone remains unchanged, our modifications introduce additional learned parameters for hierarchical depth and sibling index embeddings. Quantitatively, our introduced hierarchical embeddings comprise two additional embedding tables, each encoding depth and sibling indices. Collectively, these tables add approximately 789,504 parameters, representing roughly 0.945\% of the original model’s total 83,504,416 parameters. This minimal increase ensures that the overall complexity and computational overhead remain manageable and closely comparable to the base CodeBERTa-small model. Each experiment is conducted across three independent runs with random seeds set to 12345, 550, and 42 to ensure robust statistical evaluation. Both Masked Language Modeling (MLM) and clone detection fine-tuning were trained for three epochs using the AdamW optimizer with a learning rate of $1 \times 10^{-5}$ and a batch size of 32. Additionally, the Tree Attention Mask was selectively applied to special tokens, guiding the attention mechanism towards structurally significant tokens within the AST. Performance is measured using loss, accuracy, F1 score, precision, and recall to evaluate the impact of hierarchical embeddings.

To facilitate reproducibility, we provide an anonymized repository containing the full implementation, training scripts, and pre-processing details. See Appendix~\ref{sec:supplementary} for access.

\section{Results}

We evaluate Tree-Enhanced CodeBERTa on two tasks: masked language modeling (MLM) and clone detection. Across both, our model consistently outperforms the baseline in accuracy, F1 score, precision, and recall. Additionally, we report final training loss values to reinforce these improvements. On MLM, the Weighted Sum configuration achieves a lower loss of \textbf{0.41417} compared to \textbf{0.44388} for the original model. Similarly, in clone detection, our model attains a loss of \textbf{0.21799} versus \textbf{0.25836} for the baseline. These reductions confirm that incorporating hierarchical positional embeddings not only improves task-specific performance but also facilitates more effective representation learning.

\subsection{Masked Language Modeling (MLM)}

Table~\ref{tab:mlm-results} demonstrates consistent performance improvements from tree-based positional embeddings, particularly highlighting the Weighted Sum strategy as the most effective approach.

\begin{table}[h]
    \footnotesize
    \centering
    \begin{tabular}{lcccc} 
        \hline 
        \textbf{Embedding} & \textbf{Acc.} & \textbf{F1} & \textbf{Precision} & \textbf{Recall} \\
        \hline 
        Original        & 0.8972 & 0.8939 & 0.8953 & 0.8972 \\
        Sum             & 0.9021 & 0.8989 & 0.9004 & 0.9021 \\
        Weighted Sum    & \textbf{0.9029} & \textbf{0.8999} & \textbf{0.9012} & \textbf{0.9029} \\
        Concatenation   & 0.9026 & 0.8993 & 0.9008 & 0.9026 \\
        \hline 
    \end{tabular} 
    \caption{MLM task results on the CodeSearchNet dataset (averaged over 3 runs with different random seeds: 12345, 550, and 42). Standard deviations across seeds were consistently low ($<$ 0.002), indicating stable performance improvements.} 
    \label{tab:mlm-results}
\end{table}

\subsection{Clone Detection}

Table~\ref{tab:clone-results} highlights the improved performance of Tree-Enhanced CodeBERTa in distinguishing semantically and structurally similar code snippets, with the Weighted Sum approach consistently achieving the best overall performance.

\begin{table}[h]
    \footnotesize
    \centering
    \begin{tabular}{lcccc}
        \hline
        \textbf{Embedding} & \textbf{Acc.} & \textbf{F1} & \textbf{Precision} & \textbf{Recall} \\
        \hline 
        Original & 0.9173 & 0.9172 & 0.9180 & 0.9173 \\
        Sum & 0.9159 & 0.9159 & 0.9164 & 0.9159 \\
        Weighted Sum & \textbf{0.9187} & \textbf{0.9186} & \textbf{0.9191} & \textbf{0.9187} \\
        Concatenation & 0.9063 & 0.9063 & 0.9072 & 0.9063 \\
        \hline 
    \end{tabular}
    \caption{Average performance over 3 runs on clone detection (seeds: 12345, 550, and 42). Standard deviations across seeds were below 0.002 for accuracy and F1 scores, reflecting statistically stable gains.}
    \label{tab:clone-results}
\end{table}

\paragraph{Qualitative Analysis of Representations}

\begin{figure*}[ht]
    \centering
    \includegraphics[width=\textwidth]{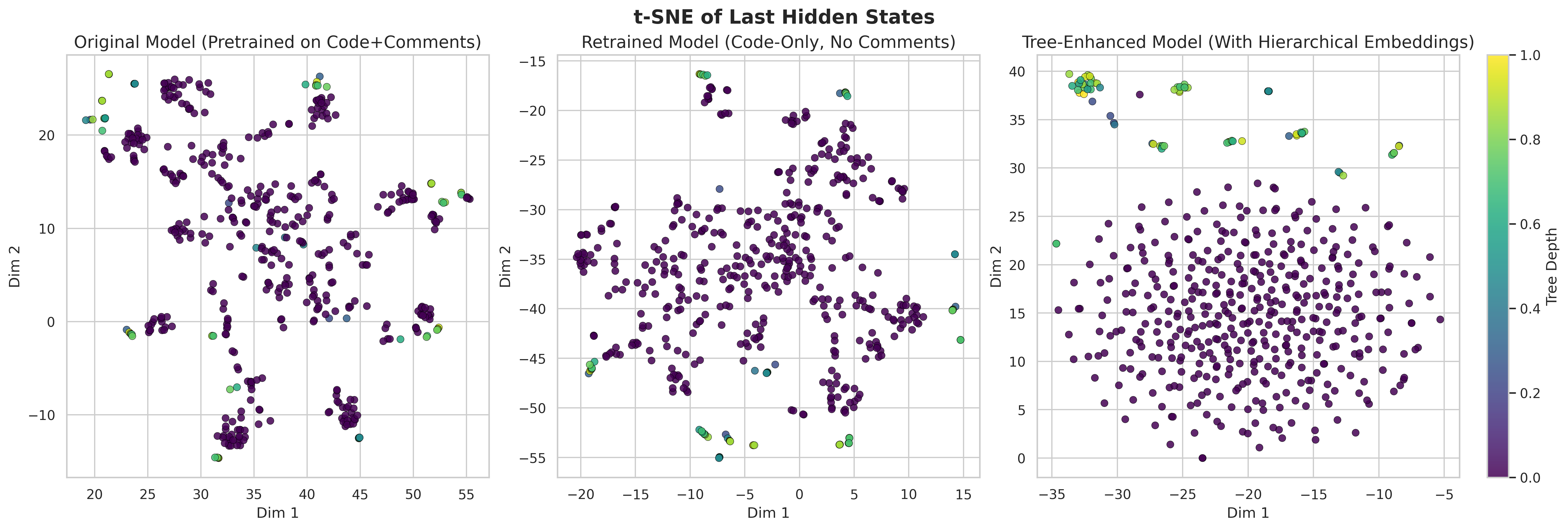}
    \caption{t-SNE projection of the last hidden states for the three models. The Tree-Enhanced Model demonstrates clearer structural clustering, indicating improved hierarchical representations.}
    \label{fig:tsne-hidden-states}
\end{figure*}

Figure~\ref{fig:tsne-hidden-states} provides a qualitative comparison of the learned representations through t-SNE projections. The visualization illustrates the hidden state embeddings from three model variants: (1) the original pretrained Transformer (trained on code and comments), (2) a retrained Transformer (trained exclusively on code without comments), and (3) our proposed Tree-Enhanced Transformer with hierarchical embeddings, using the \textbf{Weighted Sum} configuration. Each point represents an individual AST node (token), colored according to its normalized depth within the AST, with lighter colors indicating nodes situated deeper in the AST structure.

The \textbf{original pretrained model} (left) exhibits loosely formed and overlapping clusters. Nodes are grouped primarily by token-level semantic similarities without clear correlation to their structural positions within the AST hierarchy, suggesting reliance mainly on linear sequential context and token semantics from natural language comments rather than syntactic relationships.

The \textbf{retrained model} (center), trained solely on code data, demonstrates tighter clustering compared to the pretrained model due to domain specialization. However, these clusters still exhibit minimal correlation with AST depth, indicating that structural hierarchy remains underrepresented, and token representations are primarily semantic rather than structural.

In contrast, the \textbf{Tree-Enhanced model} (right) distinctly captures hierarchical structure, as evidenced by clearly delineated and depth-correlated clusters. Nodes deeper in the AST (lighter colors) form separate, well-defined peripheral clusters, while nodes nearer the AST root (darker colors) group cohesively at the center. This clear structural differentiation highlights the model’s ability to represent syntactic context explicitly, confirming the effectiveness of hierarchical embeddings.

This visualization aligns closely with our theoretical foundations, validating that the integration of tree-based positional embeddings significantly enhances the model’s capacity to encode hierarchical relationships inherent in source code, resulting in improved performance on structurally-sensitive tasks such as clone detection and masked language modeling.

\section{Analysis and Insights}

\subsection{Impact of Tree-Based Embeddings}

Our results confirm that hierarchical positional embeddings enhance the structural awareness of Transformer models for source code. By encoding hierarchical relationships, these embeddings improve representation learning, particularly in tasks that rely on structural context. In masked language modeling, they help predict contextually relevant tokens even when distant in sequence but closely related in the AST. For clone detection, they improve differentiation between structurally similar yet semantically distinct snippets, reducing false positives and boosting overall accuracy and F1 scores.

\subsection{Embedding Integration Strategies}

Our experiments highlight the trade-offs among different embedding integration methods:

\begin{itemize}
    \item \textbf{Sum Embeddings}: Computationally efficient but lacks adaptability in balancing structural and semantic contributions.
    \item \textbf{Concatenation Embeddings}: Enhances expressiveness but introduces higher dimensionality and computational cost without consistent gains.
    \item \textbf{Weighted Sum Embeddings}: Achieves the best balance, dynamically adjusting emphasis on structural embeddings, particularly in early training.
\end{itemize}

The Weighted Sum approach emerges as the most effective, offering an optimal trade-off between efficiency and structural representation quality.

\section{Limitations}

While our proposed Tree-Enhanced CodeBERTa shows significant improvements in capturing hierarchical source code structures, our approach has several inherent limitations that must be acknowledged:

\begin{enumerate}
    \item \textbf{Computational Overhead}: Integrating AST-based positional embeddings requires additional preprocessing steps, such as AST parsing and alignment, increasing computational overhead. This could limit scalability, especially in real-time or low-resource environments.
    
    \item \textbf{Parser Dependency}: Our embeddings heavily rely on the accuracy and language-specific implementation of the AST parser (Tree-Sitter \cite{tree_sitter}). Variations in parser quality or completeness across different programming languages may impact the consistency and reliability of the embeddings.
    
    \item \textbf{Generalizability Beyond Source Code}: Our method explicitly leverages hierarchical AST structures. Thus, its applicability is inherently limited to data that can be clearly represented through tree-based hierarchies. Its effectiveness on non-hierarchical or general graph structures without clear parent-child relationships remains uncertain.
\end{enumerate}

\section{Conclusion}

We introduced Tree-Enhanced CodeBERTa, a Transformer-based model incorporating hierarchical positional embeddings from Abstract Syntax Trees (ASTs). By integrating depth and sibling index embeddings, our approach captures structural nuances overlooked by traditional positional encodings.

Evaluations on masked language modeling (MLM) and clone detection confirm that these embeddings enhance representation learning, improving accuracy, F1 score, precision, and recall. The Weighted Sum integration strategy proves most effective, balancing structural and semantic information while maintaining efficiency.

\subsection{Key Takeaways}
\begin{itemize}
    \item Tree-based positional embeddings improve source code understanding by explicitly modeling hierarchical structure.
    \item The Weighted Sum integration strategy optimally balances semantic and structural embeddings with minimal overhead.
    \item Structural embeddings are particularly beneficial for tasks like clone detection, where syntactic differentiation is critical.
\end{itemize}

\subsection{Future Work}
Future directions include optimizing AST parsing for computational efficiency and exploring language-agnostic intermediate representations (IRs), such as data flow graphs, to mitigate the strict dependency on syntax rules and enhance cross-language generalization. Additionally, hierarchical embeddings could be extended beyond source code, potentially benefiting tasks involving natural language parse trees or structured document analysis.

These findings highlight the potential of hierarchical positional embeddings for structured data representation, paving the way for further exploration in broader applications.

\bibliography{bibliography}

\appendix
\appendix
\section{Supplementary Materials}
\label{sec:supplementary}

To facilitate reproducibility, we provide an anonymized repository containing the full implementation, training scripts, and pre-processing details:

\begin{center}
\url{https://anonymous.4open.science/r/tree-enhanced-codebert-BC1B}
\end{center}

This repository includes:
\begin{itemize}
    \item Source code for model training and evaluation.
    \item Scripts for preprocessing source code into ASTs using Tree-Sitter.
    \item Model hyperparameters and configuration files.
\end{itemize}

\end{document}